# Multistage Hybrid Arabic/Indian Numeral OCR System

Yasser M. Alginaih, Ph.D., P.Eng. IEEE Member
Dept. of Computer Science
Taibah University
Madinah, Kingdom of Saudi Arabia
yginahi@taibahu.edu.sa

Abdul Ahad Siddiqi, Ph.D., Member IEEE & PEC
Dept. of Computer Science
Taibah University
Madinah, Kingdom of Saudi Arabia
asiddiqi@taibah.edu.sa

*Abstract*— The use of OCR in postal services is not yet universal and there are still many countries that process mail sorting manually. Automated Arabic/Indian numeral Optical Character Recognition (OCR) systems for Postal services are being used in some countries, but still there are errors during the mail sorting process, thus causing a reduction in efficiency. The need to investigate fast and efficient recognition algorithms/systems is important so as to correctly read the postal codes from mail addresses and to eliminate any errors during the mail sorting stage. The objective of this study is to recognize printed numerical postal codes from mail addresses. The proposed system is a multistage hybrid system which consists of three different feature extraction methods, i.e., binary, zoning, and fuzzy features, and three different classifiers, i.e., Hamming Nets, Euclidean Distance, and Fuzzy Neural Network Classifiers. The proposed system, systematically compares the performance of each of these methods, and ensures that the numerals are recognized correctly. Comprehensive results provide a very high recognition rate, outperforming the other known developed methods in literature.

I. INTRODUCTION

Optical Character Recognition (OCR) is the electronic translation of images of printed or handwritten text into machine-editable text format; such images are captured through a scanner or a digital camera. The research work in OCR encompasses many different areas, such as pattern recognition, machine vision, artificial intelligence, and digital image processing. OCR has been used in many areas, e.g., postal services, banks, libraries, museums to convert historical scripts into digital formats, automatic text entry, information retrieval, etc.

The objective of this work is to develop a numerical OCR system for postal codes. Automatic Arabic/Indian numeral OCR systems for Postal services have been used in some countries, but still there are problems in such systems, stemming from the fact that machines are unable to read the crucial information needed to distribute the mail efficiently. Historically, most civilizations have different symbols that convey numerical values, but the Arabic version is the simplest and most widely acceptable. In most Middle Eastern countries both the Arabic (0,1,2,3,4,5,6,7,8,9) and Indian (٠,١,٢,٣,٤,٥,٦,٧,٨,٩) numerals are used. The objective of this work is to develop a numeral Arabic/Indian OCR system to recognize postal codes from mail letters processed in the Middle Eastern countries. A brief history on the development of postal services is qouted from [1]. "The broad development of mechanization in postal operations was not applied until the mid-1950s. The translation from mechanization to automation of the U.S. Postal Services (USPS) started in 1982, when the first optical character reader was installed in Los Angeles. The introduction of computers revolutionized the postal industry, and since then, the pace of change has accelerated dramatically [1]."

In the 1980s, the first OCRs were confined to reading the Zip Code. In the 1990s they expanded their capabilities to reading the entire address, and in 1996, the Remote Computer Reader (RCR) for the USPS could recognize about 35% of machine printed and 2% of handwritten letter mail pieces. Today, modern systems can recognize 93% of machine-printed and about 88% of handwritten letter mail. Due to this progress in recognition technology the most important factor in the efficiency of mail sorting equipment is the reduction of cost in mail processing. Therefore, a decade intensive investment in automated sorting technology, resulted in high recognition rates of machine-printed and handwritten addresses delivered by state-of- the-art systems [1 – 2]

According to the postal addressing standards [3], a standardized mail address is one that is fully spelled out and abbreviated by using the postal services standard abbreviations. The standard requires that the mail addressed to countries outside of the USA must have the address typed or printed in Roman capital letters and Arabic numerals. The complete address must include the name of addressee, house number with street address or box number/zip code, city, province, and country. Examples of postal addresses used in the Middle East are given in table 1.

TABLE 1: Examples of postal addresses used in the Middle East

| Address with Arabic numerals | Address with Indian Numerals |
|---|---|
| Mr. Ibrahim Mohammad | السيد محمد علي |
| P.O. Box 56577 | ص. ب: ٥٢١٠٦ |
| RIYADH 11564 | الرياض: ١٢٣٤٥ |
| SAUDI ARABIA | المملكة العربية السعودية |

Standards are being developed to make it easy to integrate newer technologies into available components instead of replacing such components, which is very costly; such standards are the OCR/Video Coding Systems (VCS) developed by the European Committee for standardization. The OCR/VCS enables postal operators to work with different suppliers on needed replacements or extensions of sub-systems without incurring significant engineering cost [1] [4].

Many research articles are available in the field of automation of postal systems. Several systems have been developed for address reading, such as in USA [5], UK [6], Japan [7], Canada [8], etc. But very few countries in the Middle East use automated mail-processing systems. This is due to the absence of organized mailing address systems, thus current processing is done in post offices which are limited and use only P.O. boxes. Canada Post is processing 2.8 billion letter mail pieces annually through 61 Multi-line Optical Character Readers (MLOCRs) in 17 letter sorting Centers. The MLOCR – Year 2000 has an error rate of 1.5% for machine print reading only, and the MLOCR/RCR – Year 2003 has an error rate of 1.7% which is for print/script reading. Most of these low read errors are on handwritten addresses and on outgoing foreign mail [9].

The postal automation systems, developed so far, are capable of distinguishing the city/country names, post and zip codes on handwritten machine-printed standard style envelopes. In these systems, the identification of the postal addresses is achieved by implementing an address recognition strategy that consists of a number of stages, including pre-processing, address block location, address segmentation, character recognition, and contextual post processing. The academic research in this area has provided many algorithms and techniques, which have been implemented. Many OCR systems are available in the market, which are multi font and multilingual. Moreover, most of these systems provide high recognition rate for printed characters. The recognition rate is between 95% - 100%, depending on the quality of the scanned images, fed into the systems, and the application it is used for [9]. The Kingdom of Saudi Arabia has also initiated its efforts in deploying the latest technology of automatic mail sorting. It is reported in [10], that Saudi Post has installed an advanced Postal Automation System, working with a new GEO-data based postal code system, an Automatic Letter Sorting Machine, and an OCR for simultaneous reading of Arabic and English addresses. It comprises components for automatic forwarding, sequencing, and coding

In his in-depth research study, Fujisawa, in [11] reports on the key technical developments for Kanji (Chinese character) recognition in Japan. Palumbo and Srihari [12] described a Hand Written Address Interpretation (HWAI) system, and reported a throughput rate of 12 letters per second. An Indian postal automation based on recognition of pin-code and city name, proposed by Roy et al in [13] uses Artificial Neural Networks for the classification of English and Bangla postal zip codes. In their system they used three classifiers for the recognition. The first classifier deals with 16-class problem (because of shape similarity the number is reduced from 20) for simultaneous recognition of Bangla and English numerals. The other two classifiers are for recognition of Bangla and English numerals, individually. Ming Su et al. [14], developed an OCR system, where the goal was to accomplish the automatic mail sorting of Chinese postal system by the integration of a mechanized sorting machine, computer vision, and the development of OCR. El-Emami and Usher [15] tried to recognize postal address words, after segmenting these into letters. A structural analysis method was used for selecting features of Arabic characters. On the other hand, U.Pal et.al., [16], argues that under three-language formula, the destination address block of postal document of an Indian state is generally written in three languages: English, Hindi and the State official language. Because of intermixing of these scripts in postal address writings, it is very difficult to identify the script by which a pin-code is written. In their work, they proposed a tri-lingual (English, Hindi and Bangla) 6-digit full pin-code string recognition, and obtained 99.01% reliability from their proposed system whereas error and rejection rates were 0.83% and 15.27%, respectively. In regards to recognizing the Arabic numerals, Sameh et.al. [17], described a technique for the recognition of optical off-line handwritten Arabic (Indian) numerals using Hidden Markov Models (HMM). Features that measure the image characteristics at local, intermediate, and large scales were applied. Gradient, structural, and concavity features at the sub-regions level are extracted and used as the features for the Arabic (Indian) numeral. The achieved average recognition rate reported was 99%.

Postal services are going to remain an integral part of the infrastructure for any economy. For example, recent growth in e-commerce has caused a rise in international and domestic postal parcel traffic. To sustain the role of mail as one of most efficient means of business communication, postal services have to permanently improve their organizational and technological infrastructure for mail processing and delivery [4]. Unfortunately, as explained above the character recognition process is not perfect, and errors often occur.

A simplified illustration of how an OCR system is incorporated into postal services is shown in Figure 1. This figure, in no way reflects the current technology used in available mail processing systems. Typically, an OCR system is developed for the application of postal services in order to improve the accuracy of mail sorting by recognizing the scanned Arabic and Indian numerical postal codes from addresses of mail letters.

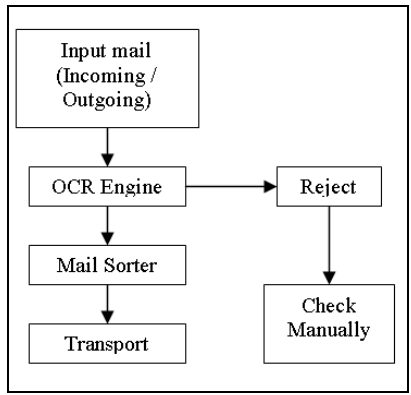

Figure 1: OCR in Postal Services

The proposed method combines different feature extraction and classification algorithms to produce a high recognition rate in such application. The proposed hybrid system is explained in section II of this paper, which explains, the different feature extraction, training and classification techniques, where as section III of this paper presents the results and observations and finally the concluding remarks are stated in section IV.

II. PROPOSED HYBRID OCR SYSTEM

The significance of this research project is in recognizing and extracting the most essential information from addresses of mail letters, i.e., postal zip codes. This system will have a profound effect in sorting mail and automating the postal services system, by reading the postal codes from letter addresses, which contain Arabic (0, 1, 2, 3…) and Indian (٠, ١, ٢, ٣, ٤….) numerals. Therefore, this system can be considered a bi-numeral recognition system.

The proposed system includes more than one feature extraction and classification methods. As a result, the hybrid system will help reduce the misclassification of numerals. The system can be used specifically in the Middle East and countries which use Arabic and Indian numerals in their documents. The proposed design methodology includes a character recognition system, which goes through different stages, starting from preprocessing, character segmentation, feature extraction and classification. The main building blocks of a general OCR system are shown in Figure 2 and the design of the proposed hybrid system is shown in Figure 3.

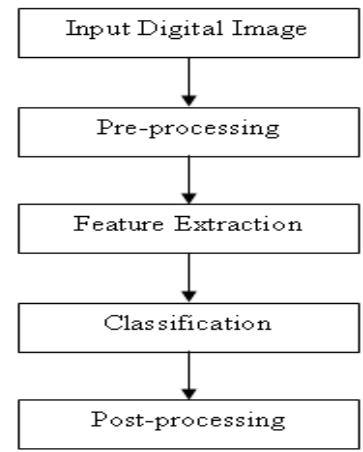

Figure 2: A General OCR System

Figure 2, represents the stages a general OCR system goes through. The process here ignores all the steps before the OCR step and assumes the availability of the mail document as a grey-level bitmap graphic file. The proposed OCR system in Figure 3 show the preprocessing, feature extraction, and classification stages. It also shows stage for comparison to produce the output recognized numeral. After the preprocessing stage, features are extracted using the first two feature extraction methods, namely feature1 and feature2, then these two feature vectors are passed through classifiers, namely classifier1 and classifier2 respectively. The output from both classifiers is compared, if there is a match then the output is accepted and no further processing is required for this numeral, otherwise the third feature is calculated, and then passed through classifier3. The output from classifier3 is then compared with both outputs of classifier1 and classifier2. If there is a match with the output of classifier3 with either outputs of classifier1 and classifier2, then the output is accepted,

otherwise the output is rejected and the postal letter needs to go through either post-processing or manual sorting.

In the next subsections of this paper, the preprocessing, feature extraction, training and classification techniques used in this system are explained in details.

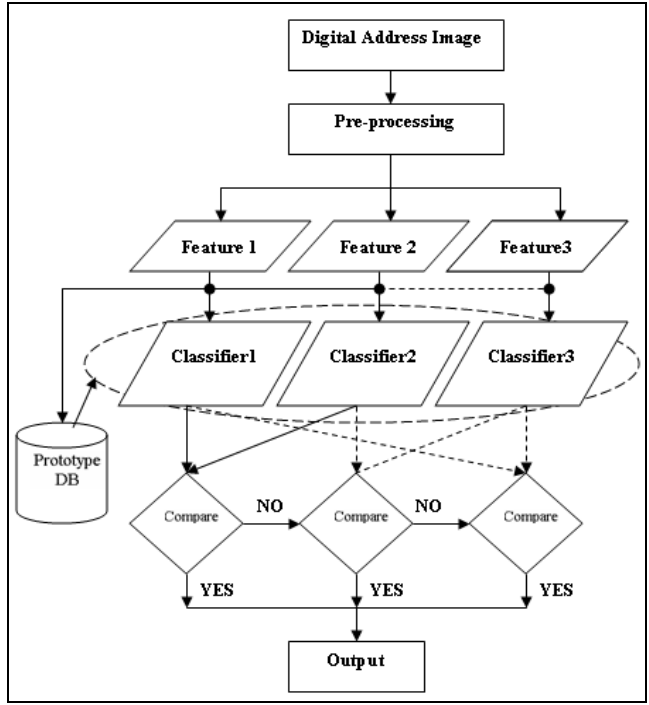

Figure 3: Proposed Hybrid OCR System

*A. Preprocessing*

Postal mail images were assumed to be free of noise with a skew angle not exceeding $\pm 2^o$. The preprocessing tasks performed are: localization of the address, conversion from grey scale images to binary images, localization of the postal code on the image, and character segmentation. The first step in pre-processing locates the address to be processed, such as the incoming/outgoing addresses, as long as the address is in the proper standard format there will not be a problem in specifying its location. Following the localization of the postal code, thresholding was used to convert the image into binary. If the pixel value was above a threshold value then it becomes white (background) otherwise black (foreground) [18]. Here, the average of the pixels in the image was taken to be the threshold value. At this stage, most of the noise was eliminated using thresholding and only slight distortion to characters was observed, which suggests that pixels were either lost or added to the characters during the thresholding process. Isolated noise was removed during the character segmentation process. Then the zip or postal code was located according to the location according to the postal services standards. After locating the postal code, the characters are segmented so that each can be processed individually for proper recognition. At this point, all numerals were normalized to a size of 25 x 20, which was decided experimentally according to a 12-font size numeral scanned at a resolution of 300 dpi. The normalization step aims to remove the variations of printed styles and obtain standardized data.

*B. Feature Extraction*

The proposed hybrid OCR system, Figure 3, is based on the feature extraction method of character recognition. Feature extraction can be considered as finding a set of vectors, which effectively represents the information content of a character. The features were selected in such a way to help in discriminating between characters. The proposed system uses a combination of three different methods of feature extraction, which are extracted from each normalized numeral in the postal code, these features are: the 2D array of the pixel values after the conversion of the address image into binary, the array of black pixel distribution values from square-windows after dividing each normalized character into a 5x5 equal size windows [19], and finally the maximized fuzzy descriptive features, [20 – 21], are obtained using equation (1).

$$S_{ij} = \max_{x=1}^{N1}(\max_{y=1}^{N2}(w[i-x, j-y]f_{xy})) \quad ---->(1)$$

$$for \quad i = 1\,to\,N_1,\, j = 1\,to\,N_2$$

$S_{ij}$ gives the maximum fuzzy membership pixel value using the fuzzy function, $w[m,n]$, equation (2). Where $f_{xy}$ is the $(x, y)$ binary pixel value of an input pattern $(0 \leq f_{xy} \leq 1)$. $N_1$ and $N_2$ are the height and width of the character window.

$$w[m,n] = \exp(-\beta^2(m^2 + n^2)) \quad ---->(2)$$

Through exhaustive search, $\beta = 0.3$ is found to be the most suitable value to achieve higher recognition rate. This maximized membership fuzzy function, equation (2), was used in the second layer of the Fuzzy Neural Network presented in [20 – 21], which will be used as one of the classifiers of the proposed system. $S_{ij}$ gives a 2D fuzzy feature vector whose values are between 0 and 1, and has the same size as the normalized image window of the numeral. It is known from the fuzzy feature vector method, that the features which resemble the shape of the character will be easily recognized due to this characteristic of the descriptive fuzzification function. Therefore, the feature values closer to the boundary of the character will result in higher fuzzy membership value. Similarly, further from the boundary of the character will result in lower fuzzy membership value.

*C. Training*

A bitmap image file containing the sets of the numbers from 0 – 9 Arabic and from ٠ – ٩ Indian was used in the training process to calculate the prototype features. The training sets from which the feature prototypes were calculated contained four different typesets, these are: Arial, Times New Roman, Lucida Console, and New Courier. Each typeset contained 20 numerals for both Arabic and Indian with 5 different font sizes (12, 14, 16, 18, 20), providing us with a total of 100 numerals for each typeset and 400 numerals for the complete training set. Figure 4 shows a full set for one typeset with all the font sizes, note the figure is not to scale.

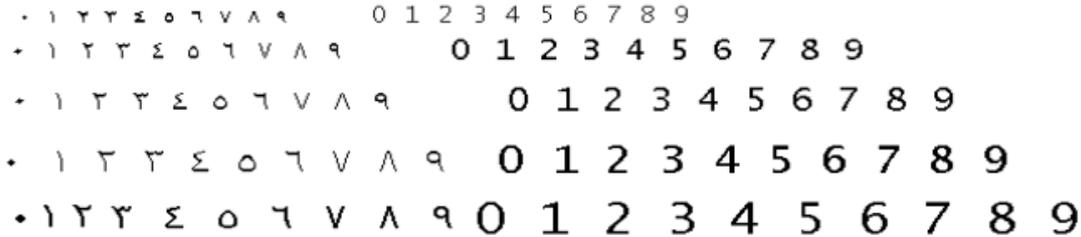

Figure 4: The training sets used to extract the prototype features. (Figure not to scale)

The prototype numerals were then normalized to a size of 25 x 20. The three different features explained above were calculated from the normalized characters, and then stored in a separate file as prototypes to be compared with the features extracted from the images under test. Figure 5(a) shows a normalized image for the Arabic numeral (1) and Figure 5(b) shows a normalized image for the Indian numeral (٤). From the image above, 1 represents the foreground and 0 represents the background of the numeral. Here, only the features for the normalized characters with font size 12 were used as the prototype features to be passed to the Hamming Net classifier since size 12 is considered as a standard size.

The prototype feature file for binary features contained 80 feature vectors, each having a vector size of 25x20 features. Figure 6 shows an example of a 32-feature vector for a normalized numeral. As explained in the feature extraction section, each of these features represents the black pixel distribution in a window size 5x5. The features from font size 12 for all sets of Arabic and Indian numerals were used as the prototype features to be passed to the Euclidean Distance Classifier. The file contained 80 vectors, each containing 32 features. Figure 6 shows an example of a 32-feature vector for an Arabic numeral.

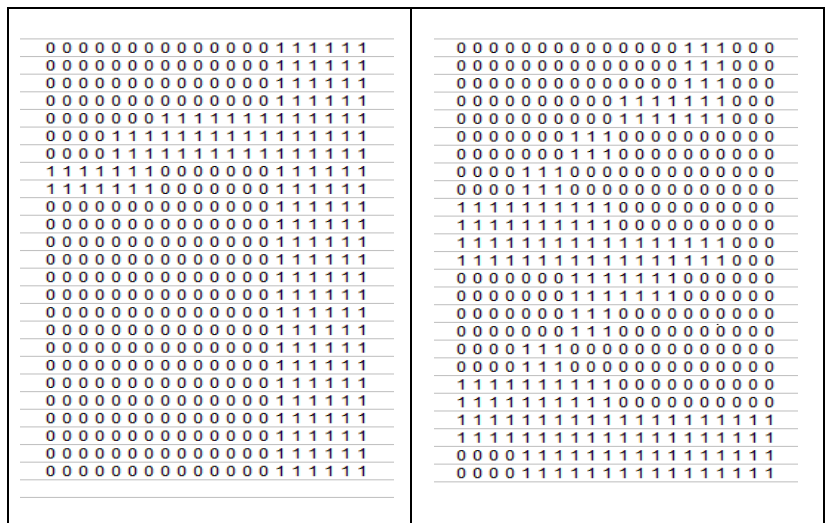

(a)                                   (b)

Figure 5: Normalized Images showing Arabic Numeral 1 and Indian Numeral 4.

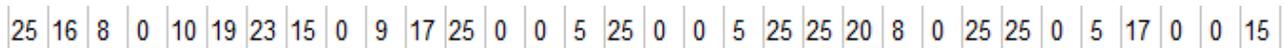

Figure 6: Zoning features for a Normalized numerals.

Figure 7 shows the fuzzy features for the Arabic numeral 1. The highlighted area resembles the shape of the numeral, which shows the fuzzy feature value equals to 1. It is from Figure 7 that, the closer to the boundary of the numeral, the higher the fuzzy feature value and, the further from the boundary of the numeral, the lower the fuzzy feature values.

![Fuzzy features table]

Figure 7: Fuzzy features for the normalized numeral 1 (Arabic) – Size 25 x 20

The prototype features were calculated from the normalized characters. For each font, the prototypes of the five font sizes for each numeral in both Arabic and Indian were averaged by adding them then dividing the sum by 5. This resulted in 20 prototype feature vectors for each typeset, 10 for Arabic numerals and 10 for Indian numerals, respectively, providing us with a total of 80 prototype feature vectors each containing 25 x 20 features as shown in Figure7. Many Arabic/Indian numeral sets for the 4 typesets were scanned at different resolutions and were used during the testing process. Figure 8 shows some examples of some numeral sets used for testing.

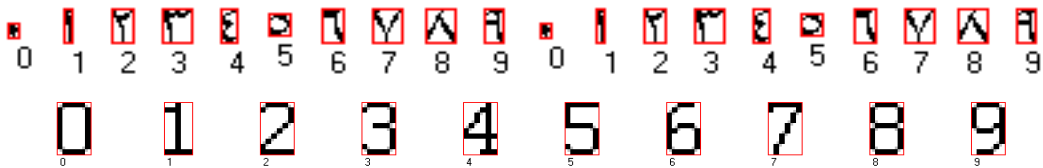

Figure 8: The results of testing some complete Arabic and Indian numeral sets

### D. Classification

A multistage OCR system with three-feature extraction and three classification algorithms is employed to maintain the accuracy in the recognition of the postal codes. The first classifier used is the Euclidean distance which provides the ordinary distance between two points. To recognize a particular input numeral feature vector, the system compares this feature vector with the feature vectors of the database of feature vectors of normalized numerals using the Euclidean distance nearest-neighbor classifier [22]. If the feature vector of the input is **q** and that of a prototype is **p**, then the Euclidean distance between the two is defined as:

$$d = \sqrt{(p_0 - q_0)^2 + (p_1 - q_1)^2 + \ldots + (p_{N-1} - q_{N-1})^2}$$

$$= \sqrt{\sum_{i=1}^{N}(p_i - q_i)^2} \quad ----------->(3)$$

Where

$p = \begin{bmatrix} p_0 & p_1 & \cdots & p_{N-1} \end{bmatrix}^T$ and
$q = \begin{bmatrix} q_0 & q_1 & \cdots & q_{N-1} \end{bmatrix}^T$

and *N* is the size of the vector containing the features. Here, the match between the two vectors is obtained by minimizing *d*.

The second classifier is the Hamming Net classifier, [23 – 24] shown in Figure 9 below.

**x** is the input vector.
O is the output of the Maxnet and it is
y is the input to the Maxnet
**c** is the encoded class prototype vector
M is the number of classes.

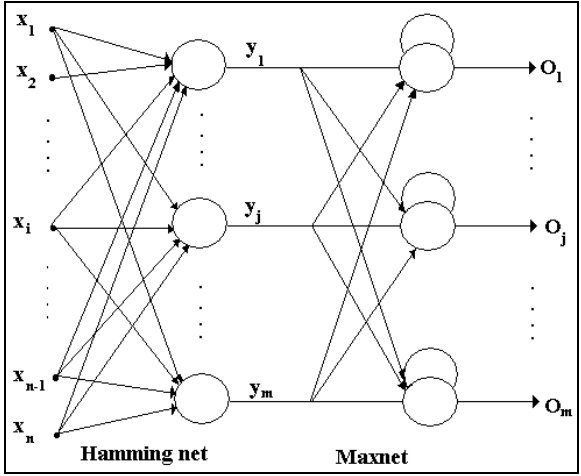

Figure 9: Hamming net with Maxnet as the second layer

The algorithm designed for the minimum Hamming distance classifier which was adopted from [23] is as follows:

Step1: initialize the weight matrix $\mathbf{w}_j$ and the biases:

$$w_{ji} \frac{c_{ji}}{2} \quad \quad ----> (4)$$

$$b_j = \frac{n}{2} \quad \quad ----> (5)$$

$$i = 1, 2, \ldots, n; \quad j = 1, 2, \ldots, M$$

Step 2: For each input vector x, do steps 3 to 5.
Step 3: Computer the $net_j$, $j = 1, 2, \ldots, M$:

$$net_j = b_j + \sum_i x_i w_{ji} \quad ----> (6)$$

$$i = 1, 2, \ldots, n; \quad j = 1, 2, \ldots, M$$

Step 4: Initialize the activation $y_j$ for the Maxnet, the second
layer of the network which represents the Hamming similarity.

$$y_j = net_j \quad \quad ----> (7)$$

Step5: Maxnet compares the outputs of the $net_j$ and enforces
the largest one as the best match prototype, while suppressing the rest to zero.
Step6: Recurrent processing of the Maxnet:

$$o = w_M y = \begin{vmatrix} 1 & -0.2 & \cdots & -0.2 \\ -0.2 & & & -0.2 \\ \vdots & & \ddots & \vdots \\ -0.2 & -0.2 & \cdots & 1 \end{vmatrix} \begin{vmatrix} y_1^k \\ y_2^k \\ \vdots \\ y_M^k \end{vmatrix} = w_M net^k \quad ---> (8)$$

Where

$$net^k = \begin{vmatrix} net_1^k \\ \vdots \\ net_j^k \\ \vdots \\ net_M^k \end{vmatrix} \quad \text{and} \quad O = \begin{vmatrix} f(net_1^k) \\ \vdots \\ f(net_j^k) \\ \vdots \\ f(net_M)^k \end{vmatrix} \quad f(net_j) = \begin{cases} 0 & \text{when } net_j < 0 \\ net_j & \text{when } net_j \geq 0 \end{cases}$$

The third classifier used in this work is the Fuzzy Neural Network, FNN, developed by Kwan and Cai, [20]. It uses the fuzzy descriptive features explained in the feature extraction section. Figure 10 shows the structure of the network which is a four-layer FNN. The first layer is the input layer; it accepts patterns into the network which consists of the 2D pixels of the input numeral. The second layer of the network is a 2D layer of MAX fuzzy neurons whose purpose is to fuzzify the input patterns through the weighted function *w[m, n]*, equation (2). The third layer produces the learned patterns. The fourth layer is the output layer which performs defuzzification and provides non-fuzzy outputs; it chooses the maximum similarity as the activation threshold to all the fuzzy neurons in the fourth layer (Refer [20 - 21] for details on the FNN). After passing through the different stages of the classifier, the character is identified and the corresponding class is assigned. In the post-processing step, recognized postal codes will be compared against valid postal codes stored in a database. In the classification phase, feature vectors of an unknown character are computed and matched with the stored prototypes. Matching is done by calculating distance (dissimilarity) measure between the character and stored prototypes.

The proposed system, shown in Figure 3 suggests that if the Hamming Net Classifier and the Euclidean Distance Classifier did not provide a match then the fuzzy features are calculated and passed through the FNN classifier. The FNN classifier is different from a traditional Neural Network because the function of each fuzzy neuron is identified and its semantics is defined. The function of such networks is the modeling of inference rules for classification. The outputs of a FNN provide a measurement of the realization of a rule, i.e. the membership of an expected class. Typically, a FNN is represented as special four-layer feed-forward neural network, in which the first layer corresponds to the input variables, the second layer symbolizes the fuzzy rules, the third layer produces the learned patterns and the fourth layer represents the output variables. It is trained by means of a data-driven learning method derived from neural network theory. Therefore, the result of the FNN classifier is compared to both other classifiers and if there is a match found between the FNN's result and any of the previously calculated classifier results the numeral is accepted, otherwise it is rejected, Figure 3.

### III. RESULTS AND OBSERVATIONS

The authors presented the initial results of this research study in [25], in which only one font was used and no thorough testing of the system was conducted. The proposed system can handle small amount of skew in the range of –2 to +2 degrees. The system supports BMP image formats; with image scan resolution of 100 – 300 dpi and above. The documents used were of multiple fonts with multiple sizes. The fonts used in the system for testing were: Arial, New Times Norman, Lucida Console and New courier. Font sizes of 10 – 20, with font styles normal, and bold were incorporated in the system.

Extensive testing of the proposed OCR system has been done on approximately 200 mail address images of different quality printed documents with different resolutions, font styles and sizes. Figure 11 shows an example of a processed mail address.

The proposed hybrid system produced successful results in recognizing Arabic and Indian numerals from postal letters. The proposed hybrid system provided a 100% recognition rate with no misclassification of numerals and a rejection rate of less than 1%. When combining the recognition rate for all images at different resolutions, the average recognition rate considering the rejected numerals as misclassified the recognition rate was 99.41% for all the images which varied in resolutions, typesets and brightness. This recognition rate assumed that the rejected characters were misclassified. This shows the effectiveness of the system in providing high recognition rate using 4 different fonts and suggests that more fonts could be applied, if desired.

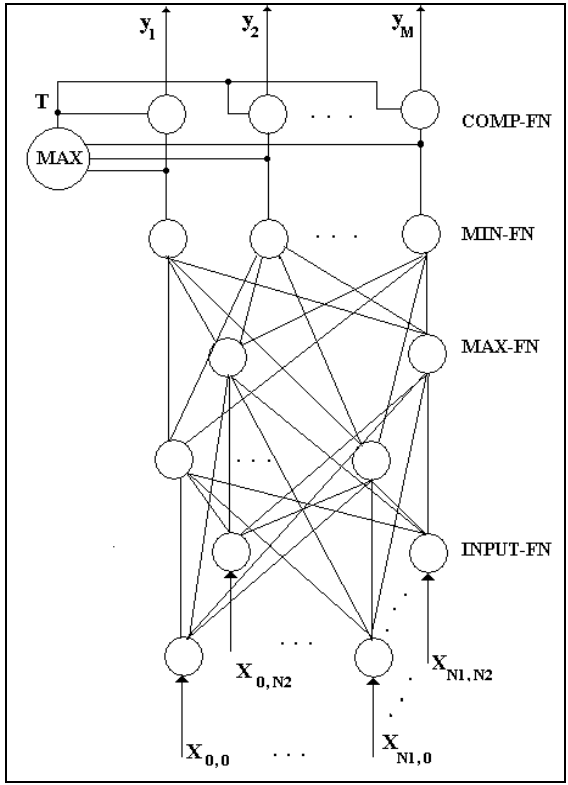

Figure 10: Four-Layer Feed forward FNN.

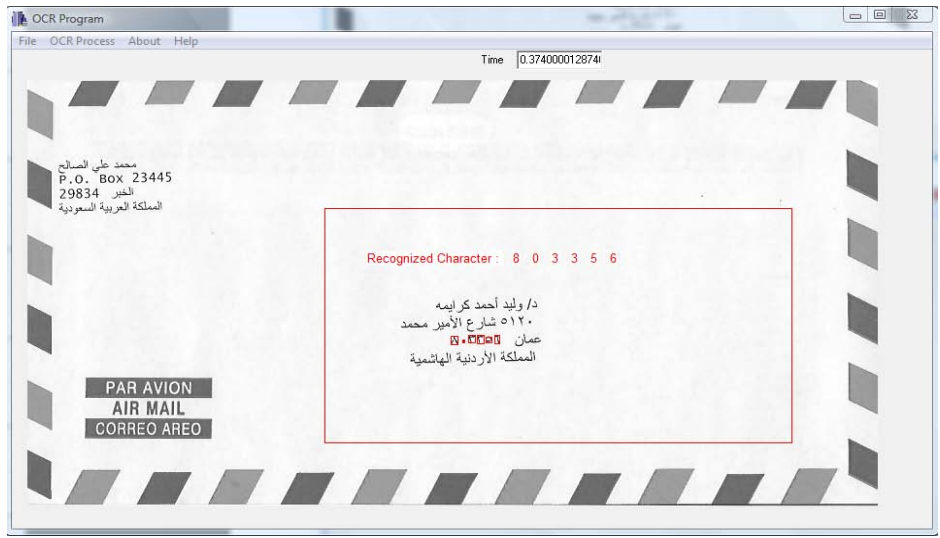

Figure 11: A processed envelope containing a postal code written in Indian numerals.

Future work will use more fonts, and will incorporate a post-processing step to check the availability of the postal codes so as to ensure the character recognition of Middle Eastern countries' addresses for mail sorting and proper distribution of mail according to postal zip codes and cities. Tables 2 – 5 show the recognition rates for the proposed hybrid method and the three methods used separately. The proposed hybrid method outperformed the other three methods, if used separately, as shown in Table 2. The recognition rate calculated in Table 1 did not include any of the rejected numerals. It can also be observed that, the higher the resolution, the better the recognition rate.

TABLE 2: Recognition rate for all methods using images with different resolutions

| | Resolution | 100% | 200% | 300% | 400% |
|---|---|---|---|---|---|
| Recognition Rate | No. of Characters | 5460 | 4340 | 5690 | 2710 |
| | Hamming | 99.08% | 99.39% | 98.80% | 98.89% |
| | Euclidean Distance | 99.36% | 99.08% | 98.76% | 99.88% |
| | Fuzzy Neural Network | 98.13% | 99.31 | 99.43% | 100% |
| | Proposed Hybrid Method | 100% | 100% | 100% | 100% |

Table 3 shows the total number of misclassified numerals at different resolutions.

TABLE 3: Number of misclassified characters using images with different resolutions

| | Resolution | 100% | 200% | 300% | 400% |
|---|---|---|---|---|---|
| Misclassified Characters | No. of Characters | 5460 | 4340 | 5690 | 2710 |
| | Hamming | 72 | 70 | 68 | 30 |
| | Euclidean Distance | 50 | 40 | 70 | 3 |
| | Fuzzy Neural Network | 146 | 30 | 32 | 0 |
| | Proposed Hybrid Method | 0 | 0 | 0 | 0 |

Table 4 shows the number of rejected characters when using the proposed hybrid method. As shown, there were no misclassified or rejected numerals with the 400% resolution - due to the fact that larger size numerals provide good quality numerals when normalized.

TABLE 4: Number of rejected characters using the proposed hybrid method

| | Proposed Hybrid Method | | | |
|---|---|---|---|---|
| Resolutions | 100% | 200% | 300% | 400% |
| No. of Rejected Characters | 40 | 30 | 38 | 0 |

Table 5 presents the recognition rate for the hybrid method when considering the rejected numerals as misclassified.

TABLE 5: Recognition rate including rejected characters for the proposed hybrid method

| | Proposed Hybrid Method | | | |
|---|---|---|---|---|
| Resolution | 100% | 200% | 300% | 400% |
| Recognition Rate | 99.27 | 99.31 | 99.33 | 100 |

IV. CONCLUSION

In this work, a hybrid numeral OCR system for Arabic/Indian postal zip codes was successfully developed and thoroughly tested. The system used three different feature extraction methods and three different classifier techniques in order to guarantee the accuracy of any numeral processed through it. Over 200 letter images were used where the postal code was localized, and then recognized through the proposed system. Four different font styles with sizes ranging from 10 to 20 points were used in testing the system and the recognition accuracy was 99.41%, when considering the rejected numerals as un-recognized numerals.

ACKNOWLEDGMENT

The authors would like to acknowledge the financial support by the Deanship of Scientific Research at Taibah University, KSA, under research reference number 429/230 academic year 2008/2009 to carry out the research to design and develop the postal OCR system for the recognition of address postal codes in the Middle Eastern countries.

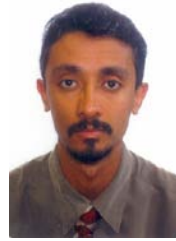

**Yasser M. Alginahi,** became a member of IEEE in 2000. He earned a Ph.D., in electrical engineering from the University of Windsor, Ontario, Canada, a Masters of Science in electrical engineering and a Bachelors of Science in biomedical engineering from Wright State University, Ohio, U.S.A. Currently, he is an Assistant Professor, Dept. of Computer Science, College of Computer Science and Engineering, Taibah University, Madinah, KSA. His current research interests are Document Analysis, Pattern Recognition (OCR), crowd management, ergonomics and wireless sensor networks. He is a licensed Professional Engineer and a member of Professional Engineers Ontario, Canada (PEO). He has over a dozen of research publications and technical reports to his credit.

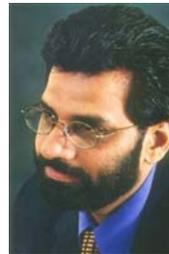

Dr. Abdul Ahad Siddiqi received a PhD and a MSc in Artificial Intelligence in year 1997, and 1992 respectively from University of Essex, U.K. He also holds a bachelor degree in Computer Systems Engineering from NED University of Engineering and Technology, Pakistan. He is a Member of IEEE, and Pakistan Engineering Council. Presently he is an Associate Professor at College of Computer Science and Engineering at Taibah University, Madinah, KSA. He has worked as Dean of Karachi Institute of Information Technology, Pakistan (affiliated with University of Huddersfield, U.K.) between 2003 and 2005. He has over 18 research publications to his credit. He has received research grants from various funding agencies, notably from Pakistan Telecom, and Deanship of Research at Taibah University for research in are areas of Intelligent Information Systems, Information Technology, and applications of Genetic Algorithms.